\documentclass[10pt,a4paper]{article}
\usepackage{fontspec}
\usepackage{array}
\newcolumntype{P}[1]{>{\centering\arraybackslash}p{#1}}
\usepackage{tabularx,array,graphicx,adjustbox,stfloats,float}
\usepackage{multirow,microtype,xcolor,url,inconsolata,makecell,hyperref,tcolorbox}
\usepackage{booktabs}

\usepackage[final]{lrec2026}

\title{VIVID: A Culturally Grounded Benchmark Exposing the Figurative Language Gap in Vietnamese NLP}

\name{\parbox{\linewidth}{\centering
Tu Tran Do$^{1*}$, Nhat Ngoc Nguyen$^{1*}$, Khanh-Tung Tran$^{2}$ \\
Hoang D. Nguyen$^{2}$,Tu Minh Phuong$^{1}$, Long Hoang Dang$^{1}$
}}
\address{
$^{1}$Posts and Telecommunications Institute of Technology (PTIT), Vietnam \\
$^{2}$School of Computer Science and Information Technology, University College Cork, Ireland \\
\texttt{\{tudt.b22kh109, nhatnn.b22at208\}@stu.ptit.edu.vn} \\
\texttt{123128577@umail.ucc.ie \quad hn@cs.ucc.ie} \\
\texttt{\{phuongtm, longdh\}@ptit.edu.vn}
}

\abstract{
We present VIVID (Vietnamese Idioms for Validation and Interpretation Depth), the first systematic benchmark for evaluating culturally grounded figurative language understanding in Vietnamese. VIVID comprises 1,636 idioms and proverbs annotated with five complexity traits (literal expressions, pragmatic nuances, Sino-Vietnamese terms, uncommon vocabulary, folk knowledge) and seven semantic themes. We establish an evaluation framework combining generative and discriminative tasks, proposing an LLM-as-a-Judge approach with aspect-based prompting validated against human judgment (Cohen's κ = 0.792). Evaluating eight state-of-the-art models reveals critical gaps: Vietnamese-specialized models drastically underperform multilingual systems (VinaLLaMA-7B: 0.13 vs. GPT-4o: 2.46), and even top models achieve less than 50\% correctness on average. Notably, few-shot prompting does not universally improve performance, with GPT-4o exhibiting degradation due to stylistic overfitting. Our analysis exposes systematic failures including literal over-interpretation, lexical gaps, and pragmatic flattening, demonstrating that current models lack cultural competence for nuanced figurative interpretation. VIVID provides an essential tool for advancing figurative language understanding in culturally rich contexts. We release codes and datasets at \url{https://github.com/ReML-AI/VIVID}.
 \\ \newline \Keywords{Vietnamese Benchmark, Figurative Language Evaluation, Idioms and Proverbs } }

\begin{document}

\maketitleabstract
\begingroup
\renewcommand\thefootnote{\textsuperscript{\normalsize *}}
\footnotetext{These authors contributed equally.}
\endgroup
\section{Introduction}

Figurative language understanding remains a critical challenge in natural language processing, requiring models to integrate linguistic knowledge, cultural context, and pragmatic reasoning  \cite{liu-etal-2022-testing}. In Vietnamese linguistics, an idiom is defined as a fixed multi-word expression whose meaning cannot be straightforwardly inferred from the literal meanings of its constituent words. A proverb, in contrast, is a short, often rhythmic sentence that encapsulates folk knowledge, life experience, or moral lessons \cite{HoangPhe:2021}. Vietnamese idioms and proverbs present unique interpretation difficulties through archaic Sino-Vietnamese vocabulary, subtle pragmatic nuances (sarcasm, irony), and folk knowledge about traditional agricultural practices. Despite Vietnamese having substantial NLP resources, including benchmarks for natural language inference \cite{Huynh2022ViNLIAV}, sentiment analysis \cite{Ho2019EmotionRF}, and multi-hop reasoning \cite{le-etal-2022-vimqa}, \textbf{no systematic evaluation framework exists for assessing culturally grounded figurative language understanding}.

This gap motivates two research questions. \textbf{First, how do state-of-the-art language models perform on Vietnamese figurative language?} Existing benchmarks focus on high-resource languages like English \cite{flute}, Korean \cite{kulture}, or Ethiopian languages \cite{proverbeval} evaluating binary idiomaticity detection rather than cultural understanding. Without a Vietnamese benchmark, we cannot assess whether Vietnamese-specialized or multilingual models accurately interpret idioms, nor identify which linguistic features challenge these systems. \textbf{Second, how can we reliably evaluate figurative language understanding?} Traditional metrics like BLEU fail to capture semantic nuances \cite{tang-etal-2024-creative, wang-etal-2025-proverbs}, while human evaluation is expensive and unscalable.

To address these questions, we introduce \textbf{VIVID} (Vietnamese Idioms for Validation and Interpretation Depth), a novel benchmark of 1,636 Vietnamese idioms and proverbs with dual-layer annotations. First, we develop a linguistically-motivated taxonomy organizing expressions into three aspects (Meaning, Lexical, Agricultural \& Customary Knowledge) with five complexity traits that capture primary interpretation challenges. Second, we categorize idioms into seven semantic themes (Criticism, Life Lessons, Work and Nature, Social Relationships, Love, Virtues, Other) reflecting Vietnamese cultural discourse. We evaluate eight state-of-the-art models—including Vietnamese-specialized systems (Vistral-7B, VinaLLaMA-7B, GreenMind-14B) and large multilingual models (GPT-4o, Gemini Flash 2.5, Llama-4-Scout, Qwen-3,Llama-SEA-LION-v3)—across complementary evaluation tasks.

For evaluation, we develop a comprehensive evaluation framework combining generative and discriminative approaches. For generative evaluation, we utilize an LLM-as-a-Judge framework where models generate idiom explanations evaluated by GPT-4.1. Through systematic comparison of four prompting strategies, we establish that aspect-based evaluation—explicitly considering factual accuracy, completeness, and native speaker alignment—achieves strongest human correlation (Cohen's κ = 0.792). For discriminative evaluation, we apply multiple-choice classification tasks for topic and linguistic characteristic identification using exact-match evaluation \cite{eval-harness}.

Our findings reveal critical gaps across all model types: Even state-of-the-art models achieve less than 50\% correctness on average, with significant variation across complexity traits—models handle Sino-Vietnamese terms better than pragmatic nuances or folk knowledge. Among comparable-sized models, Vietnamese-specialized systems (GreenMind-14B: 1.09) perform similarly to multilingual counterparts (Qwen-3-14B: 1.09), suggesting that current language-specific training does not provide advantages for cultural figurative understanding. Notably, few-shot prompting does not universally help: while benefiting most models, GPT-4o shows degradation due to stylistic overfitting. 

Our contributions are threefold: \textbf{(1)} VIVID, the first systematic benchmark for Vietnamese figurative language with 1,636 validated idiom-explanation pairs and dual-layer annotations; \textbf{(2)} evaluation protocols combining a novel LLM-as-a-Judge framework for generative assessment with discriminative classification tasks; and \textbf{(3)} comprehensive analysis showing model scale matters more than language-specific training, while exposing persistent cultural understanding challenges. VIVID provides a foundation for advancing figurative language understanding and developing culturally aware NLP systems.

\section{Related Work}
\subsection{Idiom and Proverb Datasets}

Existing figurative language datasets focus primarily on high-resource languages. \cite{flute} introduce FLUTE, a synthetic dataset of 1000 English idiomatic sentences paired with synonymous or contradictory statements. \cite{semeval} present SemEval-2022 Task 2 for multilingual idiomaticity detection across 8000 sentences in English, Portuguese, and Galician. \cite{proverbeval} compile ProverbEval with cultural proverbs from four Ethiopian languages and English, while \cite{kulture} present KorID, a cloze-test dataset of 1631 Korean idioms. These datasets advance figurative language resources but lack coverage of Vietnamese, leaving a gap in evaluating culturally grounded expressions in this language. 

\subsection{Vietnamese Benchmarks}

Vietnamese NLP has established benchmarks for natural language inference \cite{Huynh2022ViNLIAV}, sentiment analysis \cite{Ho2019EmotionRF}, hate speech detection \cite{vihos}, and comprehensive NLU evaluation through ViGLUE \cite{viglue} and VLUE \cite{viglue}. VIMQA \cite{vimqa} evaluates multi-hop reasoning capabilities. However, no dataset exists for assessing figurative and cultural language understanding in Vietnamese idioms and proverbs. VIVID addresses this gap as the first benchmark for evaluating culturally grounded figurative expressions in Vietnamese. 

\subsection{Figurative Language Evaluation}

Evaluating figurative language understanding presents unique challenges, as traditional automatic metrics like BLEU and ROUGE fail to capture semantic nuances in idiomatic interpretation \cite{tang-etal-2024-creative, wang-etal-2025-proverbs}. Recent studies demonstrate that LLM-as-a-Judge approaches \cite{Zheng2023JudgingLW} achieve higher correlation with human judgment for figurative language tasks. \cite{Rezaeimanesh2025} shows strong alignment for Persian-English idiom translation, while \cite{tang-etal-2024-creative} and \cite{proverbs},validate similar effectiveness for East Asian and Chinese idioms. However, existing work primarily establishes that LLM judges outperform automatic metrics without systematically investigating how prompting strategies affect evaluation reliability for figurative language. Different approaches—zero-shot scoring, demonstration examples, chain-of-thought reasoning, or aspect-based evaluation—may yield varying alignment with human judgment, yet this remains underexplored for culturally grounded expressions. We address this gap through systematic comparison of four prompting strategies for Vietnamese idiom evaluation, establishing that aspect-based evaluation achieves strongest human alignment (Cohen's κ = 0.792). This provides methodological guidance for reliable figurative language assessment in low-resource contexts.

\section{Dataset}
\begin{figure*}[h]
    \centering
    \includegraphics[width=1.0\textwidth]{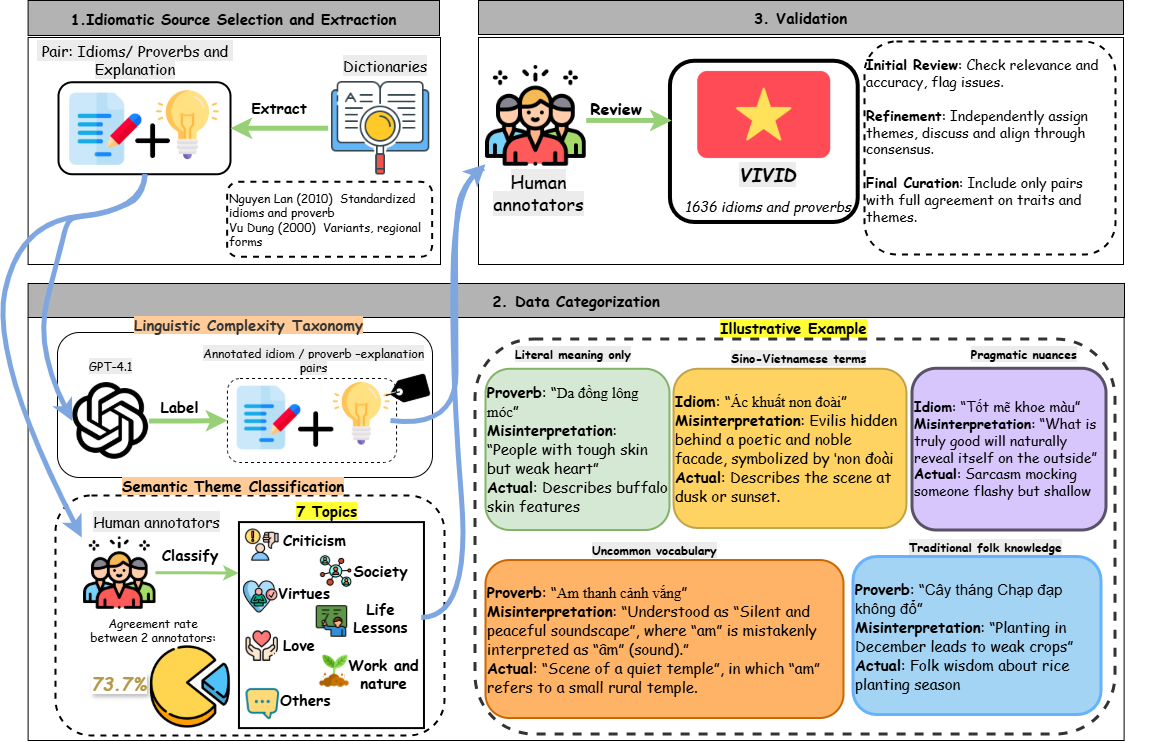}
    \caption{Overview of the data collection and annotation pipeline.}
    \label{fig:data-pipeline}
\end{figure*}

We present VIVID, a culturally grounded evaluation benchmark dataset contains 1636 idioms and proverbs across 7 topics that capture Vietnamese cultural contexts, ranging from everyday life to specific subject areas, as well as Vietnamese grammar and linguistics

\begin{figure}[h]
    \centering
    \includegraphics[width=0.5\textwidth]{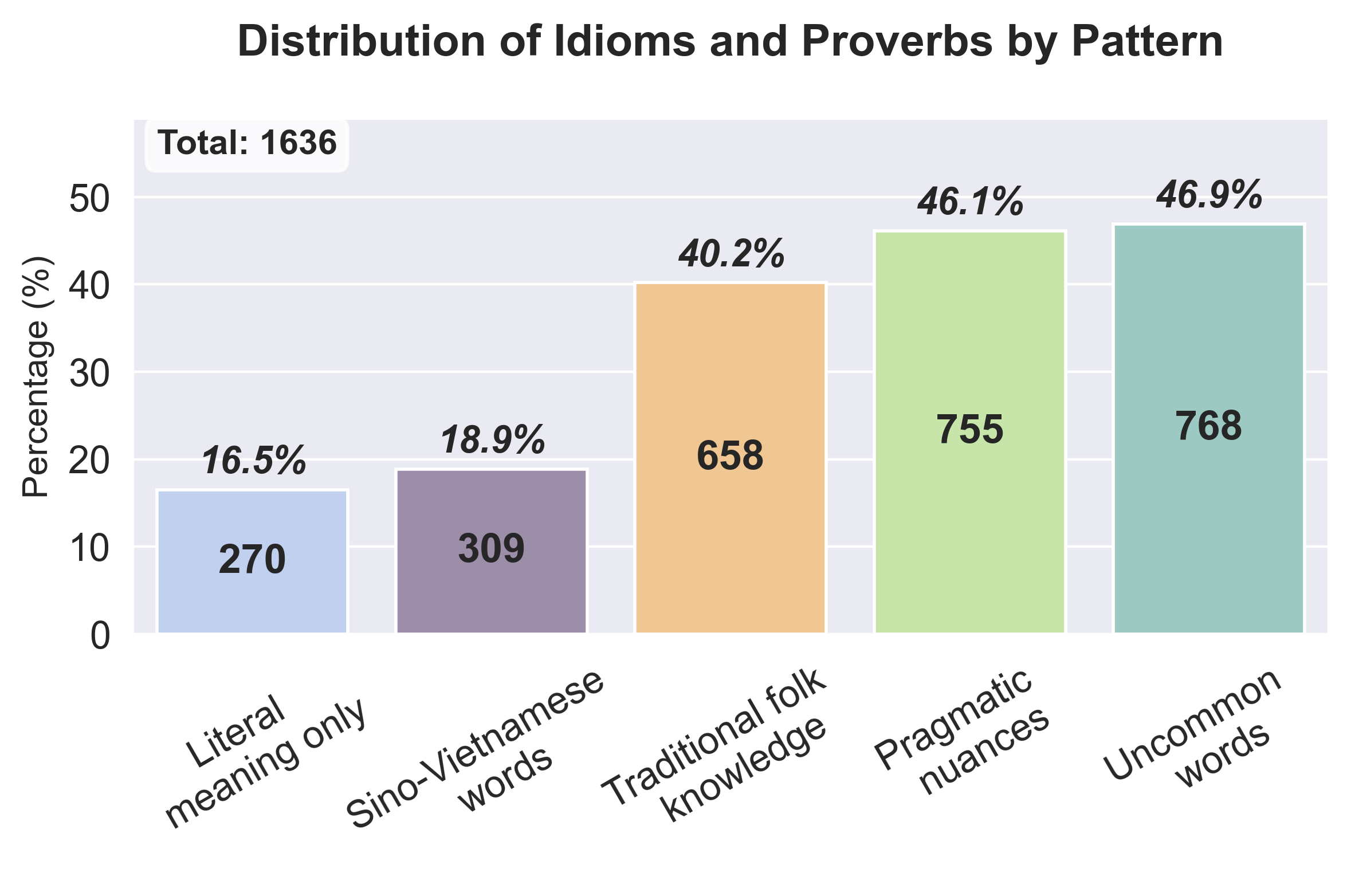}
    \caption{Statistics of VIVID per characteristics.}
    \label{fig:Statistics_chracteristics}
\end{figure}

\begin{figure}[h]
    \centering
    \includegraphics[width=0.5\textwidth]{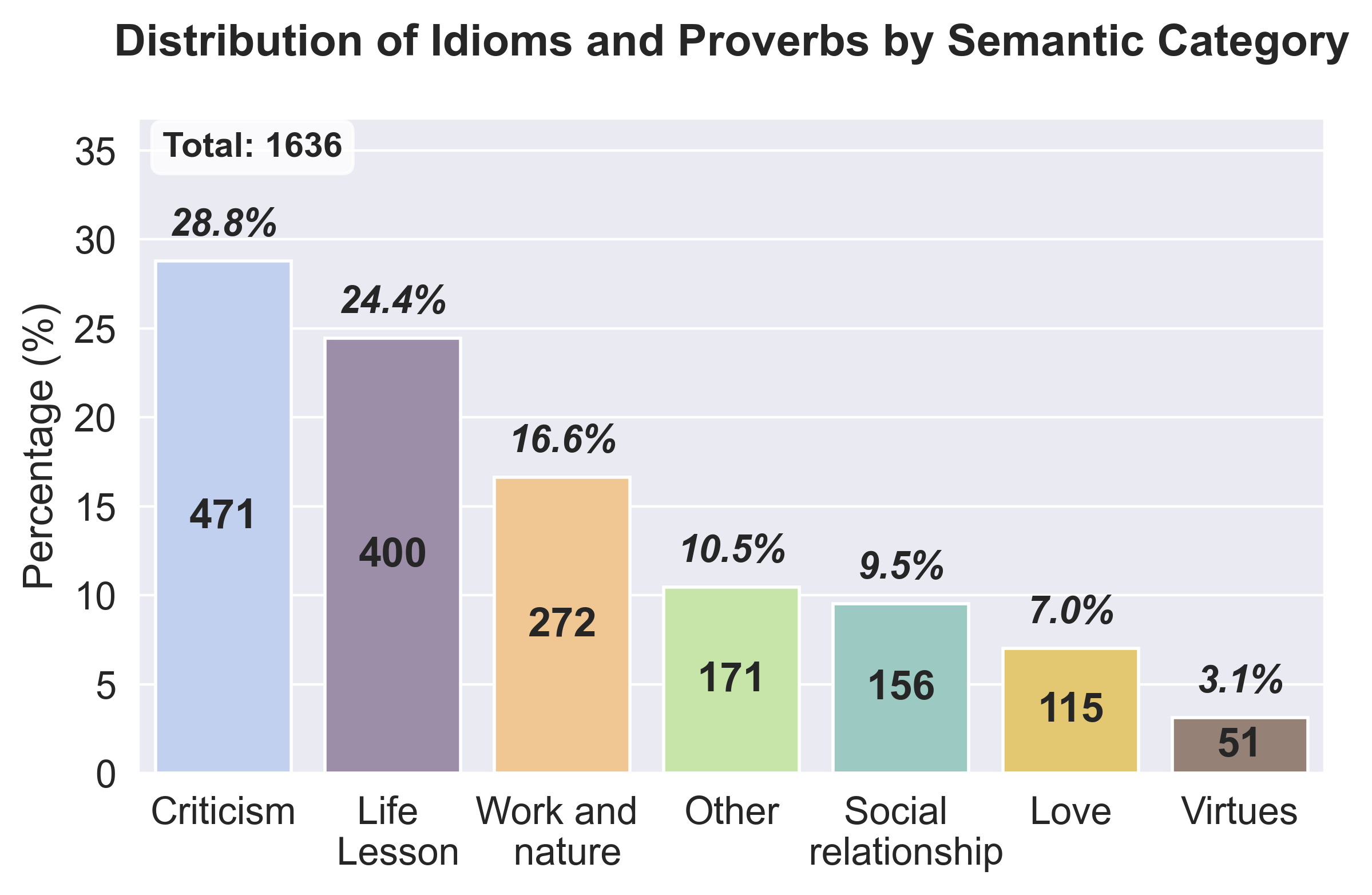}
    \caption{Statistics of VIVID per topics.}
    \label{fig:Statistics_topic}
\end{figure}

We further annotate the dataset into five distinct characteristics that LLMs are highly prone to misunderstanding. Because idioms may exhibit multiple characteristics simultaneously, the total frequency counts exceed the number of idioms. Dataset statistics are reported in Figure ~\ref{fig:Statistics_chracteristics} and ~\ref{fig:Statistics_topic}. The dataset is constructed in three stages: \ref{Data Collection} . Data Collection , \ref{Data Categorization}.  Data Categorization and \ref{Data Validation}. Data Validation, summarized in Figure \ref{fig:data-pipeline}

\subsection{Data Collection} \label{Data Collection}
To construct VIVID, we adopted a systematic approach to gather high-quality Vietnamese idioms and proverbs from authoritative lexicographic sources. Our data collection process prioritized linguistic authenticity, cultural representativeness, and semantic diversity. 

We compiled idioms and proverbs from two complementary authoritative dictionaries. The primary source was the dictionary by Nguyen Lan \cite{NguyenLan:2010}, which provides a structured and comprehensive collection of widely-used Vietnamese idioms and proverbs accompanied by clear definitions and semantic explanations. This resource is recognized as a standard reference in Vietnamese linguistics and language education, ensuring the reliability and acceptance of the selected expressions 

To enhance the linguistic diversity and capture regional variations, we incorporated additional entries from the dictionary compiled by Vu Dung and collaborators \cite{VuDung:2000}. Unlike Nguyen Lan's work, which predominantly features standardized forms, the Vu Dung dictionary documents numerous idiomatic variants including regional expressions, alternative phrasings, and lexical variations that reflect the richness of Vietnamese idiomatic usage across different dialects and contexts. This dual-source approach enables our dataset to capture both standardized expressions and their naturally occurring variants in Vietnamese communication.

Through this compilation process, we assembled a total of 2,254 idiom-explanation pairs that form the foundation of the VIVID benchmark.

\begin{figure}[h]
    \centering
    \includegraphics[width=0.5\textwidth]{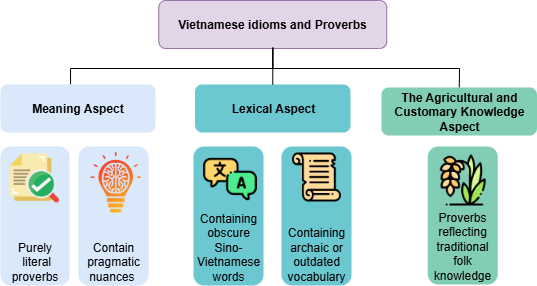}
    \caption{Taxonomy of idiomatic expressions that pose challenges for LLMs due to literal meaning, archaic vocabulary, cultural specificity, and other linguistic features.}
    \label{fig:taxonomy}
\end{figure}

\subsection{Data Categorization} \label{Data Categorization}
\subsubsection{Linguistic Complexity Taxonomy} \label{Linguistic Complexity Taxonomy }
Existing Vietnamese figurative language resources lack systematic categorization with respect to the linguistic and cultural features that challenge computational interpretation. To address this gap, we developed a theory-driven taxonomy grounded in consultation with Vietnamese language experts and informed by preliminary error analysis of LLM outputs.

Our taxonomy organizes Vietnamese idioms and proverbs along three principal aspects and five distinct traits that capture the primary sources of interpretation difficulty, as illustrated in Figure \ref{fig:taxonomy}

\textbf{The Meaning Aspect} encompasses idioms where semantic interpretation requires understanding beyond literal compositional meaning:

\begin{itemize}
    \item \textit{Literal expressions}: Proverbs that are purely descriptive without metaphorical layers, yet LLMs frequently over-interpret them by inferring non-existent figurative meanings (e.g., ``Da đồng lông móc" describes physical characteristics of buffalo skin but is often misinterpreted as a metaphor for human personality)
    \item \textit{Pragmatic nuances}: Expressions containing sarcasm, irony, or evaluative connotations that deviate from neutral interpretations (e.g., ``Tốt mẽ khoe màu" is intended to mock superficiality but is often misunderstood as promoting humility)

\end{itemize}
\textbf{
The Lexical Aspect} captures vocabulary-related challenges arising from archaic or specialized terminology:

\begin{itemize}
    \item \textit{Sino-Vietnamese terms}: Expressions containing rare or literary Sino-Vietnamese vocabulary infrequently encountered in contemporary usage (e.g., ``Ác khuất non đoài")
    \item \textit{Uncommon vocabulary}: Idioms featuring archaic or obsolete words no longer common in modern Vietnamese (e.g., ``Am thanh cảnh vắng," where ``am" refers to a small rural temple)
\end{itemize}

\textbf{The Agricultural and Customary Knowledge Aspect} includes expressions rooted in traditional Vietnamese cultural practices:

\textit{Folk knowledge-based expressions}: Proverbs reflecting agricultural practices, seasonal cycles, or customary wisdom from traditional Vietnamese life (e.g., ``Cấy tháng Chạp đạp không đổ," a saying based on historical farming experience)

Since idioms may exhibit multiple characteristics simultaneously, entries can be annotated with multiple traits. The distribution of these complexity characteristics across the dataset is presented in Figure \ref{fig:Statistics_chracteristics}, showing that pragmatic nuances (46.9\%) and uncommon vocabulary (46.1\%) are the most prevalent challenges, followed by Sino-Vietnamese terms (40.2\%), folk knowledge-based expressions (18.9\%), and literal expressions (16.5\%).

\subsubsection{Semantic Theme Classification}\label{Semantic Theme Classification }
To facilitate domain-specific evaluation and capture the cultural dimensions of Vietnamese idioms, we categorized each entry according to seven semantic themes that reflect prevalent topics in Vietnamese daily life, folk wisdom, and moral education. This classification schema was derived through thematic analysis of idiomatic content commonly encountered in colloquial usage, folk literature, and educational materials.

The seven themes are defined as follows:
\begin{itemize}
    \item \textbf{Criticism}: Expressions conveying disapproval, mockery, or critical judgment of behaviors, traits, or situations
    \item \textbf{Life Lessons}: Proverbs offering philosophical reflections, moral guidance, or practical wisdom about life
    \item \textbf{Work and Nature}: Idioms related to labor, agricultural practices, natural phenomena, or traditional livelihoods
    \item \textbf{Other}: Expressions addressing miscellaneous topics including abstract concepts, specific scenarios, or descriptions
    \item \textbf{Social Relationships}: Idioms reflecting interpersonal dynamics within society, including family ties, friendships, neighborhood interactions, and communal relationships
    \item \textbf{Love}: Expressions conveying romantic relationships, emotional bonds, or matters of the heart
    \item \textbf{Virtues}: Idioms emphasizing moral standards, ethical values, or ideal personal qualities
\end{itemize}
Figure \ref{fig:Statistics_topic} presents the distribution of idioms across these semantic categories, with Criticism (28.8\%) being the most represented, followed by Life Lessons (24.4\%), Work and Nature (16.6\%), Other (10.5\%), Social Relationships (9.5\%), Love (7.0\%), and Virtues (3.1\%).

\subsection{Data Validation} \label{Data Validation}
\label{subsec:validation}
To ensure the reliability and cultural appropriateness of our annotations, we implemented a rigorous two-stage validation process combining automatic annotation with expert human validation.
\textbf{Stage 1: Automatic Annotation.} We employed GPT-4.1 to automatically annotate each idiom-explanation pair with one or more labels corresponding to the predefined linguistic complexity traits in our taxonomy (Section 3.2.1). This automated labeling enabled systematic identification of idioms that are particularly prone to misinterpretation due to semantic complexity or cultural specificity, thereby establishing a foundation for subsequent human validation.
\textbf{Stage 2: Human Validation.} Two native Vietnamese speakers with expertise in Vietnamese linguistics independently reviewed all annotations through a three-step process:
\begin{enumerate}
    \item \textit{Initial Review}: Each annotator examined the LLM-generated complexity trait labels for contextual relevance and accuracy. Entries with clearly inaccurate or culturally inappropriate labels were flagged for removal, while uncertain cases were marked for collaborative discussion.
    \item \textit{Refinement}: For the semantic theme classification (Section 3.2.2), both annotators independently assigned each idiom-explanation pair to one of the seven predefined themes based on its core semantic content. Cases with disagreement or ambiguity were jointly reviewed, and both annotators collaboratively revised the labels to ensure alignment with the predefined taxonomy through consensus-based adjudication.
    \item \textit{Final Curation}: All disagreements were resolved through discussion between the two annotators. Only idiom-explanation pairs achieving unanimous agreement on both complexity traits and semantic themes were included in the final dataset.
\end{enumerate}
This validation process resulted in a high-quality curated dataset comprising 1,636 idiom-explanation pairs with validated annotations, providing a reliable foundation for evaluating LLMs' understanding of culturally grounded Vietnamese figurative language.

\section{Evaluation Framework and Experiment Setup}
\label{Data evaluation}

In this section, we introduce our figurative language evaluation framework, which integrates both \emph{generative} and discriminative  evaluations of Vietnamese idioms and proverbs under the LLM-as-a-Judge paradigm. This setup not only ensures reliable benchmarking but also extends prior evaluation research through prompting strategy analysis, role-based instructions, and human–LLM judge alignment studies.

\subsection{Generative Evaluation}
To evaluate generative capability, LLMs are asked to explain Vietnamese idioms and proverbs from VIVID. We adopt an open-ended question format rather than constrained multiple-choice, as prior work has shown that in multiple-choice settings, models may rely on superficial patterns rather than genuine understanding \cite{griot-etal-2025-pattern}. The quality of each generated explanation is scored using an LLM-as-a-Judge framework~\cite{gu2024survey}, with GPT-4.1 serving as an independent evaluator. We test four prompting strategies during evaluation: \textbf{(1) Zero-shot}, where the model is given only the explanation and asked to score it; \textbf{(2) Demonstration} \cite{brown2020language}, where several annotated examples of high- and low-quality explanations are provided beforehand; \textbf{(3) Chain-of-Thought (CoT)}  \cite{wei2022chain}, in which the model is prompted to reason step-by-step before producing a score; and \textbf{(4) Aspect-based Evaluation} (see Figure \ref{fig:prompt_judge}), where the model is instructed to explicitly consider multiple criteria such as factual accuracy, completeness and alignment with common native speaker interpretations.

Each explanation is rated on a scale from 0 to 5 based on its correctness, informativeness, and cultural appropriateness. These scores are then aggregated to compare performance across models and prompting strategies. Prior studies have shown that automatic metrics such as BLEU or COMET fail to capture the nuances of figurative language \cite{tang-etal-2024-creative,Rezaeimanesh2025,proverbs}. In contrast, GPT-based evaluations demonstrate high correlation with human judgments, particularly in figurative interpretation tasks \cite{yang2025evaluating}.

\begin{table}[ht]

\begin{tabular}{c r r}
\hline
\textbf{Prompt} & \textbf{Mean} & \textbf{Cohen's Kappa} \\
\hline
\textbf{Human Baseline} & 3.47 & 1.000 \\
Zero-shot & 3.40 & 0.774 \\
CoT & 3.70 & 0.786 \\
Demonstration & 3.44 & 0.783 \\
Aspect-based & 3.49 & 0.792 \\
\hline
\end{tabular}
\caption{Evaluation of different LLM-as-a-judge prompting strategies compared to human.}
\label{tab:evaluation_scores}
\end{table}

\subsubsection{LLM Judge Reliability Analysis}
\label{sec:automatic-evaluation}
To examine the reliability of the LLM-as-a-Judge approach, we conduct a controlled experiment on 200 Vietnamese idioms and proverbs randomly sampled from our dataset, VIVID. For each entry, GPT-4o generates an explanation under a Zero-shot setting. These outputs are then independently evaluated by GPT-4.1, using each of the four prompting strategies outlined above. Each explanation is rated on three dimensions: semantic accuracy, interpretive completeness, and alignment with native speaker understanding, using a 0–5 scale.

We collect human annotations following the procedure of \citet{Rezaeimanesh2025} for comparison with LLM judge. Two native Vietnamese speakers independently annotate 150 idiomatic expressions each, with a 50-sample overlap. Human scores are assigned on the same scale (0–5), achieving strong inter-human annotator reliability. 

As reported in Table~\ref{tab:evaluation_scores}, Aspect-based Evaluation achieves the strongest alignment with human ($\kappa = 0.792$), yielding the closest mean rating (3.49 to 3.47). Consequently, we adapt Aspect-based as the default prompting strategy for automatic scoring of figurative language generative task.

\subsection{Discriminative Evaluation}
We also conduct multiple-choice classification tasks focusing on (1) topic identification and (2) linguistic characteristic classification, using the prompts illustrated in Figure~\ref{fig:prompt_topic} and Figure~\ref{fig:prompt_pattern}. These tasks are implemented using the Language Model Evaluation Harness framework \cite{eval-harness}. To ensure a consistent and reproducible evaluation protocol across both opened and closed source models, we adopt exact-match scoring. This choice is motivated by the current limitation of certain closed models (e.g., GPT and Gemini) that do not expose log-likelihoods. All models are evaluated under a 3-shot prompt to ensure understanding of task format and output consistency.

\subsection{Benchmark Models}
We evaluate a diverse set of open-source and closed-source large language models (LLMs) that vary in scale, language specialization, and accessibility. The open-source models include Llama-4-Scout, Llama-SEA-LION-v3-8B-IT, GreenMind-Medium-14B-R1, Vistral-7B-Chat, VinaLLaMA-7B-Chat, and Qwen-3-14B, while the closed-source models include GPT-4o and Gemini 2.5 Flash (see Table~\ref{tab:model-selection} for details).

\begin{table}[t]
\centering
\begin{tabular}{p{1.5cm}|p{5.0cm}}
\hline
\textbf{Type} & \textbf{Model} \\
\hline
API-only LLMs & GPT-4o, Gemini 2.5 Flash \\
\hline
Open-source LLMs & 
\textbf{Multilingual:}

 Llama-4-Scout (109B) 
 
 Llama-SEA-LION-v3-8B-IT (8B)
 
 Qwen-3 (14B)

\par
\textbf{Vietnamese-specialized:} 
GreenMind-Medium-14B-R1 (14B) \cite{Tung2025}

Vistral-7B-Chat (7B) \cite{Vo2024}

VinaLLaMA-7B (7B) \cite{VinaLLaMA}
\\
\hline
\end{tabular}
\caption{Model selection. The numbers in parentheses indicate the parameters for models.}
\label{tab:model-selection}
\end{table}

Two prompting strategies are employed: Zero-shot (Figure \ref{fig:prompt_zeroshot}) and Few-shot prompting \cite{li2024from}. For Few-shot prompting  (Figure \ref{fig:prompt_fewshot}), three examples are provided per task, consisting of two proverbs and one idiom, selected to maximize coverage of the linguistic and cultural characteristics discussed above. This design helps models better infer the task format and expected outputs.

\section{Result \& Discussion}



\subsection{Performance on Generative Task}
\begin{table*}[t]
\centering
\begin{adjustbox}{max width=\textwidth}
\begin{tabular}{c|l|l|c|c|c|c|c|c|c|c}
\hline
\textbf{Category} & \textbf{Model} & \textbf{Prompt} & \textbf{Love} & \textbf{Virtues} & \textbf{Criticism} & \textbf{Work and nature} & \textbf{Society} & \textbf{Life Lessons} & \textbf{Others} & \textbf{Average} \\
\hline
\multirow{12}{*}{\shortstack{\textbf{Open-source}\\\textbf{LLMs}}} 
    & \multirow{2}{*}{Vistral-7B-Chat} 
        & Zero-shot & 0.25 & 0.49 & 0.24 & 0.34 & 0.42 & 0.44 & 0.39 & 0.35 \\
    &   & Few-shot  & 0.54 & 1.37 & 0.58 & 0.69 & 0.69 & 0.83 & 0.70 & 0.70 \\
\cline{2-11}
    & \multirow{2}{*}{GreenMind-Medium-14B-R1}
        & Zero-shot & 1.05 & 1.40 & 0.99 & 0.90 & 1.04 & 1.37 & 1.06 &  1.09\\
    &   & Few-shot  & 0.95 & 1.12 & 0.94 & 0.91 & 0.86 & 1.28 & 0.97 & 1.02  \\
\cline{2-11}
    & \multirow{2}{*}{Vinallama-7b-chat}
        & Zero-shot & 0.13 & 0.16 & 0.08 & 0.11 & 0.10 & 0.22 & 0.10 & 0.13 \\
    &   & Few-shot  & 0.23 & 0.33 & 0.14 & 0.16 & 0.31 & 0.41 & 0.17 & 0.24  \\
\cline{2-11}
    & \multirow{2}{*}{Llama-4-Scout-17B-16E}
        & Zero-shot & 0.86 & 1.27 & 0.92 & 0.76 & 0.88 & 0.96 & 1.01 & 0.91 \\
    &   & Few-shot  & 1.04 & 1.61 & \color{blue}{1.17} & \color{blue}{1.19} & \color{blue}{1.24} & \color{blue}{1.39} & \color{blue}{1.19} & 1.24 \\
\cline{2-11}
    & \multirow{2}{*}{Llama-SEA-LION-v3-8B-IT}
        & Zero-shot & 0.46 & 0.80 & 0.54 & 0.54 & 0.50 & 0.72 & 0.55 & 0.58 \\
    &   & Few-shot  & 0.62 & 0.61 & 0.60 & 0.68 & 0.56 & 0.78 & 0.59 & 0.65 \\
\cline{2-11}
    & \multirow{2}{*}{Qwen-3-14B}
        & Zero-shot & \color{blue}{1.15} &\color{blue}{1.75}& 0.95 & 1.01 & 1.07 & 1.28 & 0.95 & 1.09 \\
    &   & Few-shot  & 0.86 & 1.69 & 1.01 & 1.07 & 1.07 & 1.37 & 0.99 & 1.12 \\
\hline
\multirow{4}{*}{\shortstack{\textbf{API-only}\\\textbf{LLMs}}}
    & \multirow{2}{*}{Gemini Flash 2.5}
        & Zero-shot & 2.51 & 2.94 & 2.22 & 1.76 & 2.36 & 2.69 & 2.03 & 2.29 \\
    &   & Few-shot  & \textbf{2.54} & 2.86 & \textbf{2.51} & \textbf{2.44} & 2.52 & \textbf{2.89} &\textbf{ 2.46} & 2.6 \\
\cline{2-11}
    & \multirow{2}{*}{GPT-4o}
        & Zero-shot & 2.33 & \textbf{3.08} & 2.32 & 2.41 & \textbf{2.58} & 2.60 & 2.44 & 2.46 \\
    &   & Few-shot  & 1.75 & 2.90 & 1.95 & 1.91 & 1.89 & 2.31 & 1.97 & 2.04 \\
\hline
\end{tabular}
\end{adjustbox}

\caption{Mean score of the models by category (rows: model and prompt style; columns: topic and average). The highest score for each category is in bold. The
 top-performing open-source models are marked in blue.}
\label{tab:model_performance}
\end{table*}
\begin{table*}[t]
\centering

\begin{adjustbox}{max width=\textwidth}
\begin{tabular}{c|l|l|
                P{2.2cm}|   
                P{2.6cm}|
                P{2.6cm}|
                P{3.5cm}|
                P{2.6cm}}
\hline
\textbf{Category} & \textbf{Model} & \textbf{Prompt} & \textbf{Only literal expressions} & \textbf{Sino-Vietnamese terms} & \textbf{Uncommon vocabulary} & \textbf{Folk knowledge-based expressions} & \textbf{Pragmatic nuances} \\
\hline
\multirow{12}{*}{\shortstack{\textbf{Open-source}\\\textbf{LLMs}}} 
  & \multirow{2}{*}{Vistral-7B-Chat}
    & Zero-shot & 0.40 & 0.38 & 0.27 & 0.34 & 0.24 \\
  & & Few-shot  & 0.76 & 0.82 & 0.57 & 0.74 & 0.58 \\
\cline{2-8}
  & \multirow{2}{*}{GreenMind-Medium-14B-R1}
    & Zero-shot & 0.91 & 1.30 & 0.91 & 1.04 & 0.97 \\
  & & Few-shot  & 0.95 & 1.20 & 0.80 & 1.06 & 0.94 \\
\cline{2-8}
  & \multirow{2}{*}{Vinallama-7b-chat}
    & Zero-shot & 0.07 & 0.12 & 0.06 & 0.15 & 0.10 \\
  & & Few-shot  & 0.12 & 0.28 & 0.18 & 0.26 & 0.18 \\
\cline{2-8}
  & \multirow{2}{*}{Llama-4-Scout-17B-16E}
    & Zero-shot & 0.94 & 1.18 & 0.75 & 0.83 & 0.83 \\
  & & Few-shot  & \textcolor{blue}{1.19} & \textcolor{blue}{1.50} & \textcolor{blue}{1.10} & \textcolor{blue}{1.25} & \textcolor{blue}{1.18} \\
\cline{2-8}
  & \multirow{2}{*}{Llama-SEA-LION-v3-8B-IT}
    & Zero-shot & 0.48 & 0.74 & 0.46 & 0.58 & 0.54 \\
  & & Few-shot  & 0.59 & 0.79 & 0.52 & 0.70 & 0.58 \\
\cline{2-8}
  & \multirow{2}{*}{Qwen-3-14B}
    & Zero-shot & 0.85 & 1.37 & 0.92 & 1.10 & 1.00 \\
  & & Few-shot  & 0.96 & 1.34 & 0.92 & 1.16 & 1.03 \\
\hline
\multirow{4}{*}{\shortstack{\textbf{API-only}\\\textbf{LLMs}}}
  & \multirow{2}{*}{Gemini Flash 2.5}
    & Zero-shot & 1.70 & 3.04 & 2.30 & 2.15 & 2.19 \\
  & & Few-shot  & \textbf{2.34} & \textbf{3.11} & \textbf{2.45} & \textbf{2.57} & \textbf{2.55} \\
\cline{2-8}
  & \multirow{2}{*}{GPT-4o}
    & Zero-shot & \textbf{2.43} & 2.86 & 2.33 & 2.39 & 2.44 \\
  & & Few-shot  & 1.97 & 2.36 & 1.89 & 2.04 & 1.92 \\
\hline
\end{tabular}
\end{adjustbox}
\caption{Model performance across five types of linguistic and cultural complexity.}
\label{tab:linguistic_categories_performance}
\end{table*}

We evaluate eight LLMs on the idiom/proverb explanation generation task in VIVID, under Zero-shot and Few-shot prompting across seven semantic themes (Table~\ref{tab:model_performance}). Smaller models (7-8B parameters) struggle significantly, with VinaLLaMA-7B-Chat achieving only 0.13 and Vistral-7B-Chat reaching 0.35. Medium-sized models (14B parameters) perform better but remain below 1.5, with both Vietnamese-specialized GreenMind-14B (1.09) and multilingual Qwen-3-14B (1.09) showing comparable performance. This suggests that current language-specific training does not provide clear advantages for cultural figurative understanding at similar model scales. Large-scale models demonstrate substantially higher scores, with GPT-4o (2.46) and Gemini Flash 2.5 (2.29) outperforming all open-source systems, likely due to their extensive pretraining and significantly larger parameter counts. 

Nevertheless, all models struggle with idioms/proverbs (best approach with GPT-4o and zero-shot prompting achieves less than 50\% correctness on average), underscoring persistent challenges in Vietnamese figurative understanding.

\subsubsection{Robustness To Linguistic Complexity}
\label{sec: Evaluating_Model_cross_patterns}
We analyze model robustness across five annotated traits:
(1) literal expressions, (2) uncommon vocabulary, (3) rare Sino-Vietnamese terms, (4) folk knowledge–based expressions, and (5) pragmatic nuances.

Table~\ref{tab:linguistic_categories_performance} presents the average performance per category, with the Sino-Vietnamese terms category as the least challenging one for LLMs. GPT-4o and Gemini Flash 2.5 outperform all others across these dimensions. GPT-4o achieves 2.86 on Sino-Vietnamese terms and 2.44 on pragmatic nuances (Zero-shot), while Gemini Flash 2.5 (Few-shot) reaches 2.57 on folk knowledge and 2.45 on rare vocabulary. Conversely, lighter models such as VinaLLaMA-7B and Vistral-7B remain below 1.0, highlighting their limited semantic and cultural generalization capacity.

Few-shot prompting generally improves performance across most categories, confirming the benefit of in-context examples for linguistically complex content. The largest gain is seen in Llama-4-Scout-17B ($1.18 \rightarrow 1.50$ on Sino-Vietnamese terms). Notably, GPT-4o shows the opposite trend: its scores drop slightly in Few-shot, suggesting overfitting or semantic rigidity when exposed to structured exemplars. We further analyze this anomaly in Section~\ref{sec:Investigating_Prompting}.

Manual inspection reveals systematic failures: literal over-interpretation (imposing figurative readings where none exist), lexical gaps (mistranslating archaic vocabulary), cultural disconnection (losing contextual meaning in folk-based idioms), and pragmatic flattening (replacing sarcasm or irony with neutral tone). These patterns expose fundamental gaps in current models' cultural competency.

\subsubsection{Investigating Prompting Failures in GPT-4o on VIVID}
\label{sec:Investigating_Prompting}

GPT-4o uniquely shows consistent performance decline from zero-shot to few-shot prompting. Analysis reveals GPT-4o overfits to exemplar patterns, imitating moralizing tone rather than focusing on semantic fidelity. For example, ``Chưa lại người" (physical recovery after illness) was correctly interpreted in zero-shot but reinterpreted in few-shot as metaphor for immaturity. This indicates didactic examples induce stylistic mimicry and semantic drift, fabricating moral lessons absent in source idioms.

Few-shot prompting also degrades vocabulary handling. ``Lang đuôi thì bán, lang trán thì cày" (about buffalo coat patterns) was correctly interpreted under zero-shot but misread under few-shot, where ``lang" was mistaken for ``sweet potato," producing an unrelated agricultural metaphor. Similar findings by \cite{phelps-etal-2024-sign} suggest few-shot prompting can hinder figurative interpretation by encouraging stylistic imitation over contextual understanding.

\subsection{Evaluating Topic and Linguistic Characteristic Classification Tasks}

\begin{table*}[t]
\centering
\begin{adjustbox}{max width=\textwidth}
\renewcommand{\arraystretch}{1.05} 
\begin{tabular}{l|l|c|c}
\hline
\textbf{Category} & \textbf{Model} & \textbf{Topic} & \textbf{Linguistics and cultural} \\
\hline
\multirow{6}{*}{\textbf{Open-source LLMs}} 
  & Llama-SEA-LION-v3-8B-IT & 0.319 & 0.013 \\
  
  & GreenMind-Medium-14B-R1 & 0.455 & 0.035 \\
  
  & Qwen-3-14B & 0.447 & 0.031 \\

  & Llama-4-Scout-17B-16E & \textcolor{blue}{0.459} & \textcolor{blue}{0.042} \\
  
  & Vinallama-7B-Chat & 0.079 & 0.016 \\
  
  & Vistral-7B-Chat & 0.000 & 0.000 \\
\hline
\multirow{2}{*}{\textbf{API-based LLMs}}
  & GPT-4o & 0.524 & \textbf{0.184} \\
  
  & Gemini Flash 2.5 & \textbf{0.535} & 0.144 \\
\hline
\end{tabular}
\end{adjustbox}
\caption{Model accuracy across topic and linguistic-cultural classification tasks.}
\label{tab:topic_linguistics_labeling}
\end{table*}

We evaluate model performance on two discriminative tasks: topic classification and linguistic–cultural taxonomy classification. Accuracy results are summarized in Table~\ref{tab:topic_linguistics_labeling}.

Open-source LLMs show considerable variation across both tasks, while API-based models consistently outperform them, likely due to larger parameter scales and broader pretraining. Among open-weight systems, Llama-4-Scout-17B-16E achieves the highest scores in both topic and linguistic–cultural labeling (0.459 and 0.042), followed by Qwen3-14B (0.447, 0.031) and GreenMind-Medium-14B-R1 (0.455, 0.035). Nevertheless, accuracy on linguistic characteristic classification remains low overall, partly because the exact-match metric penalizes even minor deviations (e.g., added or omitted words). For instance, Viet-Mistral-7B-Chat scored zero in both tasks due to inconsistent adherence to output format and limited capacity (7B parameters). Closed-source models such as GPT-4o and Gemini 2.5 Flash perform best, achieving 0.524 and 0.535 on topic classification, and 0.184 and 0.144 on linguistic–cultural labelling, respectively.

Performance on linguistic–cultural characteristic classification is much lower than on topic classification, reflecting the complexity of culturally grounded expressions. Errors often arise from difficulty interpreting idioms’ implicit semantics. For example, ``ăn tái ăn tam'' (``to eat rare, to eat three times''), literally describing repetitive eating, belongs to the \emph{Other} category but was misclassified by both GPT-4o and Gemini 2.5 Flash as \emph{Criticism}, likely due to reliance on affective cues. Similarly, ``bánh giầy nếp cái, con gái họ'' (``glutinous rice cake, their daughter''), a proverb metaphorically praising a beautiful girl (\emph{Virtues}), was misclassified by GPT-4o as \emph{Love} and by Gemini 2.5 Flash as Other. These cases illustrate the persistent gap between surface-level lexical cues and the culturally embedded semantics required for accurate classification.

These results demonstrate that state-of-the-art  LLMs still struggle to fully capture the meaning and cultural dimensions of Vietnamese idioms and proverbs, underscoring the value of VIVID as a benchmark for advancing culturally aware NLP evaluation.

\section{Conclusion}

We present VIVID, the first systematic benchmark for evaluating culturally grounded figurative language understanding in Vietnamese, comprising 1,636 idioms and proverbs annotated with linguistic complexity traits and semantic themes. Our extensive evaluation reveals that even state-of-the-art models achieve less than 50\% correctness on average, underscoring fundamental gaps in cultural language understanding. Our analysis reveals systematic failure modes including literal over-interpretation, lexical gaps with archaic vocabulary, cultural disconnection in folk-based expressions, and pragmatic flattening. Notably, few-shot prompting does not universally improve performance, with GPT-4o exhibiting stylistic overfitting that degrades semantic fidelity. These findings demonstrate that current LLMs lack the cultural competency required for nuanced figurative interpretation despite their general linguistic capabilities. VIVID establishes a robust evaluation framework for Vietnamese figurative language comprehension and provides actionable insights for developing culturally aware NLP systems that go beyond surface-level pattern matching to achieve genuine figurative language understanding.

\section{Limitations}
First, our automatic evaluation framework relies on a proprietary closed-source model (GPT-4.1). Dependence on a closed model raises concerns regarding transparency and long-term reproducibility. Commercial models may be updated or modified over time without full disclosure of technical details, potentially leading to changes in evaluation behavior in future versions. This may affect the stability of results if experiments are replicated at different points in time. Second, the reliability analysis between the LLM judge and human annotators is conducted on a subset of 200 samples due to the substantial cost and effort associated with manual annotation. While this sample size is adequate for estimating agreement trends between evaluators, it may not fully capture the complete diversity of linguistic forms, rarity levels, and cultural nuances present in VIVID. Future work may extend validation to a larger scale and compare multiple evaluation models, including open-source alternatives, to further enhance the transparency and reproducibility of the proposed evaluation framework.

\section{Acknowledgements}
    This publication has emanated from research supported in part by grants from Research Ireland under Grant [12-RC-2289-P2] and [18/CRT/6223] which is co-funded under the European Regional Development Fund. For the purpose of Open Access, the author has applied a CC BY public copyright licence to any Author Accepted Manuscript version arising from this submission.

\nocite{*}
\section{Bibliographical References}\label{sec:reference}

\bibliographystyle{lrec2026-natbib}
\bibliography{lrec2026-example}


\appendix

\section{Prompts}
\label{app:prompts}

\subsection{Generative Task Prompts}
\label{app:prompts_generative}

We present the exact prompt templates used for explanation generation.
All prompts are issued in Vietnamese to ensure the models respond in the
target language and leverage their Vietnamese-language capabilities.

\begin{figure}[h]
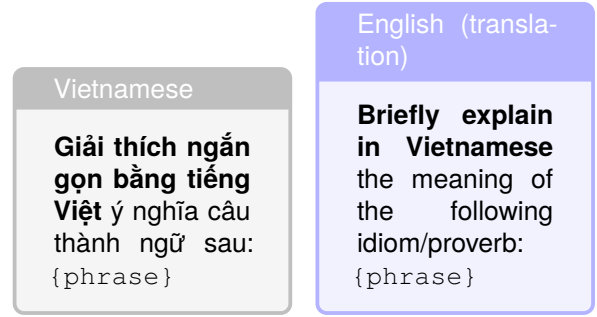

\noindent
\begin{minipage}[t]{0.48\linewidth}
\begin{tcolorbox}[colback=gray!8, colframe=gray!50, title=Vietnamese]
\textbf{Giải thích ngắn gọn bằng tiếng Việt} ý nghĩa câu thành ngữ sau: \texttt{\{phrase\}}
\end{tcolorbox}
\end{minipage}
\hfill
\begin{minipage}[t]{0.48\linewidth}
\begin{tcolorbox}[colback=blue!5, colframe=blue!30, title=English (translation)]
\textbf{Briefly explain in Vietnamese} the meaning of the following idiom/proverb: \texttt{\{phrase\}}
\end{tcolorbox}
\end{minipage}
\caption{Zero-shot prompt for idiom explanation generation (Vietnamese original left, English translation right).}
\label{fig:prompt_zeroshot}
\end{figure}

\paragraph{Zero-shot Prompt.}
The zero-shot prompt (Figure~\ref{fig:prompt_zeroshot}) is intentionally minimal, providing only the idiom or
proverb and asking for a concise Vietnamese explanation.

\begin{figure*}[h]
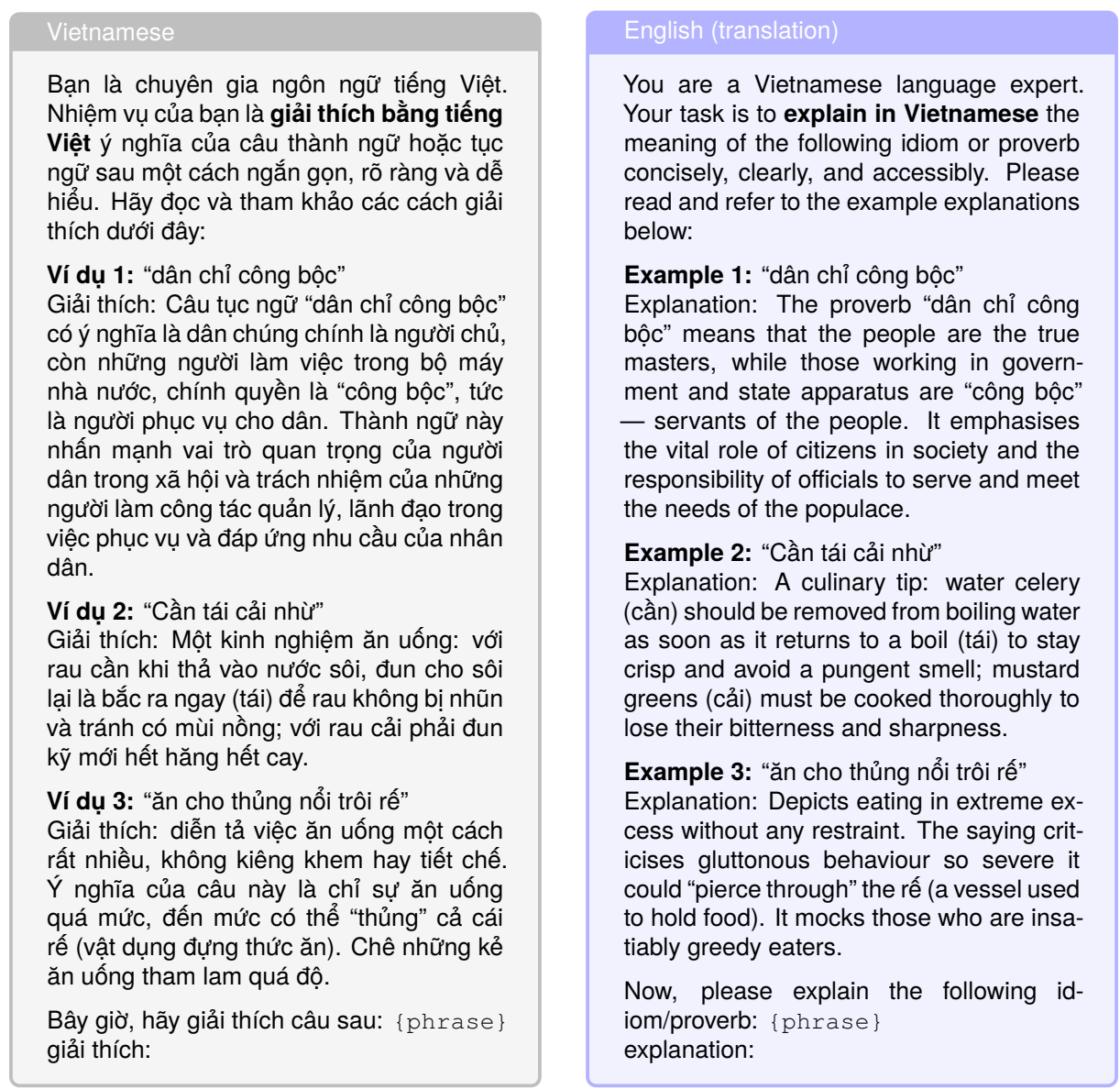

\noindent
\begin{minipage}[t]{0.48\linewidth}
\begin{tcolorbox}[colback=gray!8, colframe=gray!50, title=Vietnamese]
Bạn là chuyên gia ngôn ngữ tiếng Việt. Nhiệm vụ của bạn là \textbf{giải thích bằng tiếng Việt} ý nghĩa của câu thành ngữ hoặc tục ngữ sau một cách ngắn gọn, rõ ràng và dễ hiểu. Hãy đọc và tham khảo các cách giải thích dưới đây:

\medskip
\noindent\textbf{Ví dụ 1:} ``dân chỉ công bộc''

\noindent Giải thích: Câu tục ngữ ``dân chỉ công bộc'' có ý nghĩa là dân chúng chính là người chủ, còn những người làm việc trong bộ máy nhà nước, chính quyền là ``công bộc'', tức là người phục vụ cho dân. Thành ngữ này nhấn mạnh vai trò quan trọng của người dân trong xã hội và trách nhiệm của những người làm công tác quản lý, lãnh đạo trong việc phục vụ và đáp ứng nhu cầu của nhân dân.

\medskip
\noindent\textbf{Ví dụ 2:} ``Cần tái cải nhừ''

\noindent Giải thích: Một kinh nghiệm ăn uống: với rau cần khi thả vào nước sôi, đun cho sôi lại là bắc ra ngay (tái) để rau không bị nhũn và tránh có mùi nồng; với rau cải phải đun kỹ mới hết hăng hết cay.

\medskip
\noindent\textbf{Ví dụ 3:} ``ăn cho thủng nổi trôi rế''

\noindent Giải thích: diễn tả việc ăn uống một cách rất nhiều, không kiêng khem hay tiết chế. Ý nghĩa của câu này là chỉ sự ăn uống quá mức, đến mức có thể ``thủng'' cả cái rế (vật dụng đựng thức ăn). Chê những kẻ ăn uống tham lam quá độ.

\medskip
Bây giờ, hãy giải thích câu sau: \texttt{\{phrase\}}

\noindent giải thích:
\end{tcolorbox}
\end{minipage}
\hfill
\begin{minipage}[t]{0.48\linewidth}
\begin{tcolorbox}[colback=blue!5, colframe=blue!30, title=English (translation)]
You are a Vietnamese language expert. Your task is to \textbf{explain in Vietnamese} the meaning of the following idiom or proverb concisely, clearly, and accessibly. Please read and refer to the example explanations below:

\medskip
\noindent\textbf{Example 1:} ``dân chỉ công bộc''

\noindent Explanation: The proverb ``dân chỉ công bộc'' means that the people are the true masters, while those working in government and state apparatus are ``công bộc'' — servants of the people. It emphasises the vital role of citizens in society and the responsibility of officials to serve and meet the needs of the populace.

\medskip
\noindent\textbf{Example 2:} ``Cần tái cải nhừ''

\noindent Explanation: A culinary tip: water celery (cần) should be removed from boiling water as soon as it returns to a boil (tái) to stay crisp and avoid a pungent smell; mustard greens (cải) must be cooked thoroughly to lose their bitterness and sharpness.

\medskip
\noindent\textbf{Example 3:} ``ăn cho thủng nổi trôi rế''

\noindent Explanation: Depicts eating in extreme excess without any restraint. The saying criticises gluttonous behaviour so severe it could ``pierce through'' the rế (a vessel used to hold food). It mocks those who are insatiably greedy eaters.

\medskip
Now, please explain the following idiom/proverb: \texttt{\{phrase\}}

\noindent explanation:
\end{tcolorbox}
\end{minipage}
\caption{Few-shot prompt for idiom explanation generation, with three in-context demonstrations (Vietnamese original left, English translation right).}
\label{fig:prompt_fewshot}
\end{figure*}

\paragraph{Few-shot Prompt.}
The few-shot prompt (Figure~\ref{fig:prompt_fewshot}) situates the model as a Vietnamese language expert
and provides three demonstrative examples covering diverse idiom types
(a social proverb, a culinary proverb, and an ironic proverb) before
asking for the target explanation.

\subsection{LLM-as-a-Judge Prompt}
\label{app:prompts_judge}

The aspect-based judge is shown in Figure~\ref{fig:prompt_judge}. It instructs GPT-4.1 to evaluate a generated explanation across
four explicit dimensions before producing a single overall similarity
score on a 0--5 scale.  Crucially, a score of 0 on Criterion~1
(core-meaning accuracy) forces an overall score of 0 regardless of other
criteria, preventing superficially fluent but semantically incorrect
responses from receiving inflated scores.

\begin{figure*}[h]
\noindent
\begin{minipage}[t]{0.48\linewidth}
\begin{tcolorbox}[colback=gray!8, colframe=gray!50, title=Vietnamese]
\textbf{[Hướng dẫn]}

Bạn là chuyên gia ngôn ngữ. Nhiệm vụ của bạn là đánh giá độ tương đồng giữa giải thích gốc (con người) và giải thích của LLM cho một câu thành ngữ/tục ngữ, dựa trên các tiêu chí sau:

\medskip
\noindent 1. \textbf{Độ chính xác của ý nghĩa cốt lõi (0--5 điểm):} Giải thích của LLM nắm bắt đúng và chính xác ý nghĩa chính của câu thành ngữ/tục ngữ so với giải thích gốc? (5: Rất chính xác, 0: Sai lệch hoàn toàn)

\noindent 2. \textbf{Khả năng nắm bắt sắc thái, ngữ cảnh, ẩn ý (0--5 điểm):} Giải thích của LLM có phản ánh được các sắc thái, ngữ cảnh sử dụng điển hình, hoặc ý nghĩa ẩn sâu của câu nói như giải thích gốc không? (5: Rất tương đồng, 0: Không phản ánh được)

\noindent 3. \textbf{Chất lượng diễn đạt, sự rõ ràng, lưu loát (0--5 điểm):} Giải thích của LLM có rõ ràng, dễ hiểu, mạch lạc và tự nhiên không? (5: Rất tốt, 0: Rất kém/khó hiểu)

\noindent 4. \textbf{Độ đầy đủ (0--5 điểm):} Giải thích của LLM có bao gồm các ý chính được đề cập trong giải thích gốc không? (5: Rất đầy đủ, 0: Thiếu nhiều ý chính)

\noindent 5. \textbf{Độ tương đồng tổng thể (0--5 điểm):} Tổng hợp các đánh giá trên, mức độ tương đồng chung là bao nhiêu? (5: Rất tương đồng, 0: Không tương đồng)

\medskip
\textit{[In-context examples — see Figure~\ref{fig:judge_examples}]}

\medskip
\textbf{[Nhiệm vụ cần đánh giá]}

Thành ngữ/Tục ngữ: \texttt{\{phrase\}}

Giải thích của Con người: \texttt{\{ground\_truth\}}

Giải thích của LLM: \texttt{\{llm\_output\}}

\medskip
\medskip
Đánh giá chi tiết: \\
1. Độ chính xác của ý nghĩa cốt lõi: \\
2. Khả năng nắm bắt sắc thái, ngữ cảnh, ẩn ý: \\
3. Chất lượng diễn đạt, sự rõ ràng, lưu loát: \\
4. Độ đầy đủ: \\
Nếu tiêu chí 1 được 0 điểm thì Độ tương đồng tổng thể được 0. \\
Độ tương đồng tổng thể:
hãy suy nghĩ sau đó chỉ trả về duy nhất đánh giá của độ tương đồng tổng thể và không giải thích gì thêm
\medskip
\end{tcolorbox}
\end{minipage}
\hfill
\begin{minipage}[t]{0.48\linewidth}
\begin{tcolorbox}[colback=blue!5, colframe=blue!30, title=English (translation)]
\textbf{[Instructions]}

You are a language expert. Your task is to evaluate the similarity between the original (human-written) explanation and the LLM-generated explanation for a given Vietnamese idiom or proverb, based on the following criteria:

\medskip
\noindent 1. \textbf{Accuracy of core meaning (0--5 points):} Does the LLM explanation correctly and precisely capture the main meaning of the idiom/proverb compared to the original explanation? (5: Very accurate, 0: Completely incorrect)

\noindent 2. \textbf{Capture of nuance, context, and implicit meaning (0--5 points):} Does the LLM explanation reflect the nuances, typical usage contexts, or deeper implied meanings of the expression as in the original explanation? (5: Highly similar, 0: Does not reflect them at all)

\noindent 3. \textbf{Expression quality, clarity, and fluency (0--5 points):} Is the LLM explanation clear, easy to understand, coherent, and natural? (5: Excellent, 0: Very poor/difficult to understand)

\noindent 4. \textbf{Completeness (0--5 points):} Does the LLM explanation include the key points mentioned in the original explanation? (5: Very complete, 0: Missing many key points)

\noindent 5. \textbf{Overall similarity (0--5 points):} Synthesising the above assessments, what is the overall degree of similarity? (5: Highly similar, 0: Not similar at all)

\medskip
\textit{[In-context examples — see Figure~\ref{fig:judge_examples}]}

\medskip
\textbf{[Task to evaluate]}

Idiom/Proverb: \texttt{\{phrase\}}

Human explanation: \texttt{\{ground\_truth\}}

LLM explanation: \texttt{\{llm\_output\}}

\medskip
\medskip
Detailed evaluation: \\
1. Accuracy of core meaning: \\
2. Capture of nuance, context, and implicit meaning: \\
3. Expression quality, clarity, and fluency: \\
4. Completeness: \\
If criterion 1 receives 0 points, the overall similarity score must also be 0. \\
Overall similarity:
Think carefully, then return only the overall similarity score and provide no further explanation.
\medskip
\end{tcolorbox}
\end{minipage}
\caption{Aspect-based LLM-as-a-Judge prompt (Vietnamese original left, English translation right).
         GPT-4.1 outputs only the integer overall similarity score (0--5).}
\label{fig:prompt_judge}
\end{figure*}

\subsubsection{Judge Prompt In-Context Examples}
\label{app:judge_examples}

Two annotated examples are embedded in the judge prompt (see
Figure~\ref{fig:judge_examples}) to calibrate
the scoring scale: one high-quality response (score 5) and one
incorrect response (score 1).

\begin{figure*}[h]
\noindent
\begin{minipage}[t]{0.48\linewidth}
\begin{tcolorbox}[colback=gray!8, colframe=gray!50, title=Vietnamese]

\noindent\textbf{Ví dụ Minh họa 1} 

\noindent Thành ngữ/Tục ngữ: \textit{Chưa đủ lông đủ cánh}

\noindent Giải thích nghĩa gốc: Còn non nớt, chưa đủ khả năng tự lập.

\noindent Giải nghĩa do LLM: Câu thành ngữ ``Chưa đủ lông đủ cánh'' được dùng để chỉ một người hoặc một sự vật, sự việc nào đó chưa phát triển hoàn thiện, chưa đủ trưởng thành hoặc chưa có đủ khả năng, kinh nghiệm để thực hiện một nhiệm vụ, công việc nào đó một cách độc lập.

\noindent Đánh giá chi tiết: \\
1.  Độ chính xác của ý nghĩa cốt lõi: 5/5 \\
2.  Khả năng nắm bắt sắc thái, ngữ cảnh, ẩn ý: 5/5 \\
3.  Chất lượng diễn đạt, sự rõ ràng, lưu loát: 5/5 \\
4.  Độ đầy đủ: 5/5 \\
Độ tương đồng tổng thể: 5/5

\bigskip
\noindent\textbf{Ví dụ Minh họa 2} 

\noindent Thành ngữ/Tục ngữ: \textit{Kim chỉ có đầu}

\noindent Giải thích nghĩa gốc: Phải có trên dưới, đầu đuôi, trật tự văn phép.

\noindent Giải nghĩa do LLM: Câu thành ngữ ``Kim chỉ có đầu'' mang ý nghĩa rằng mọi việc đều có nguyên nhân và nguồn gốc, không có gì tự nhiên mà xảy ra.

\noindent Đánh giá chi tiết: \\
1.  Độ chính xác của ý nghĩa cốt lõi: 1/5 \\
2.  Khả năng nắm bắt sắc thái, ngữ cảnh, ẩn ý: 1/5 \\
3.  Chất lượng diễn đạt, sự rõ ràng, lưu loát: 3/5 \\
4.  Độ đầy đủ: 1/5 \\
Độ tương đồng tổng thể: 1/5
\end{tcolorbox}
\end{minipage}
\hfill
\begin{minipage}[t]{0.48\linewidth}
\begin{tcolorbox}[colback=blue!5, colframe=blue!30, title=English (translation)]

\noindent\textbf{Illustrative Example 1} 

\noindent Idiom/Proverb: \textit{Chưa đủ lông đủ cánh} (``Not yet fully feathered'')

\noindent Human explanation: Still inexperienced; not yet capable of independence.

\noindent LLM explanation: The idiom ``Chưa đủ lông đủ cánh'' is used to describe a person or thing that has not yet fully developed, is not sufficiently mature, or lacks the ability and experience to perform a task or role independently.

\noindent Detailed scores: \\
1. Accuracy of core meaning: 5/5 \\
2. Capture of nuance, context, and implicit meaning: 5/5 \\
3. Expression quality, clarity, and fluency: 5/5 \\
4. Completeness: 5/5 \\
Overall similarity: 5/5

\bigskip
\noindent\textbf{Illustrative Example 2} (low-quality response)

\noindent Idiom/Proverb: \textit{Kim chỉ có đầu} (``The needle and thread have a head'')

\noindent Human explanation: There must be a proper order — top and bottom, beginning and end, social hierarchy and etiquette.

\noindent LLM explanation: The idiom ``Kim chỉ có đầu'' means that everything has a cause and origin; nothing happens spontaneously.

\noindent Detailed scores: \\
1. Accuracy of core meaning: 1/5 \\
2. Capture of nuance, context, and implicit meaning: 1/5 \\
3. Expression quality, clarity, and fluency: 3/5 \\
4. Completeness: 1/5 \\
Overall similarity: 1/5
\end{tcolorbox}
\end{minipage}
\caption{In-context scoring examples embedded in the judge prompt to calibrate the 0--5 scale
         (Vietnamese original left, English translation right).}
\label{fig:judge_examples}
\end{figure*}

\section{Discriminative Task Prompts}
\label{app:disc_prompts}

The discriminative evaluation is implemented via the Language Model
Evaluation Harness framework \citep{eval-harness} using exact-match
scoring under 3-shot prompting.  All prompts are in Vietnamese.

\subsection{Topic Classification Prompt}
\label{app:disc_topic}

The topic classification prompt is shown in Figure~\ref{fig:prompt_topic}. Models are asked to identify which of seven semantic themes best
describes a given idiom or proverb, returning a single integer (1--7).

\begin{figure*}[h]
\noindent
\begin{minipage}[t]{0.48\linewidth}
\begin{tcolorbox}[colback=gray!8, colframe=gray!50, title=Vietnamese]
Bạn là chuyên gia ngôn ngữ. Dưới đây là một câu thành ngữ/tục ngữ: ``\texttt{\{phrase\}}''

Nhiệm vụ của bạn là xác định câu này thuộc một trong các đặc điểm sau:

\medskip
\noindent 1. \textbf{Tình yêu đôi lứa và hôn nhân} \\
\quad Ví dụ: \textit{chẳng nên tình trước nghĩa sau, có con ta gả cho nhau thiệt gì}

\noindent 2. \textbf{Ca ngợi phẩm chất và đạo đức con người} \\
\quad Ví dụ: \textit{cân quắc anh hùng}

\noindent 3. \textbf{Phê phán và châm biếm các thói quen xấu, hành vi tiêu cực} \\
\quad Ví dụ: \textit{ai biết uốn câu cho vừa miệng cá}

\noindent 4. \textbf{Kinh nghiệm sống về sản xuất, lao động và thiên nhiên} \\
\quad Ví dụ: \textit{ác tắm thì ráo, sáo tắm thì mưa}

\noindent 5. \textbf{Các mối quan hệ xã hội, gia đình} \\
\quad Ví dụ: \textit{ắm con chồng hơn bồng cháu ngoại}

\noindent 6. \textbf{Triết lý và bài học về cách sống} \\
\quad Ví dụ: \textit{con ruồi đậu nặng đồng cân}

\noindent 7. \textbf{Các chủ đề khác} (nếu không phù hợp với các mục trên)

\medskip
Chỉ trả lời duy nhất một số (1--7). Không giải thích, không thêm ký tự nào khác.
\end{tcolorbox}
\end{minipage}
\hfill
\begin{minipage}[t]{0.48\linewidth}
\begin{tcolorbox}[colback=blue!5, colframe=blue!30, title=English (translation)]
You are a language expert. Below is a Vietnamese idiom or proverb: ``\texttt{\{phrase\}}''

Your task is to identify which of the following categories best describes it:

\medskip
\noindent 1. \textbf{Romantic love and marriage} \\
\quad Example: \textit{chẳng nên tình trước nghĩa sau, có con ta gả cho nhau thiệt gì}

\noindent 2. \textbf{Praising human virtues and moral character} \\
\quad Example: \textit{cân quắc anh hùng}

\noindent 3. \textbf{Criticising and satirising bad habits or negative behaviour} \\
\quad Example: \textit{ai biết uốn câu cho vừa miệng cá}

\noindent 4. \textbf{Life experience about production, labour, and nature} \\
\quad Example: \textit{ác tắm thì ráo, sáo tắm thì mưa}

\noindent 5. \textbf{Social and family relationships} \\
\quad Example: \textit{ắm con chồng hơn bồng cháu ngoại}

\noindent 6. \textbf{Philosophy and life lessons} \\
\quad Example: \textit{con ruồi đậu nặng đồng cân}

\noindent 7. \textbf{Other topics} (if none of the above apply)

\medskip
Reply with a single number only (1--7). No explanation, no extra characters.
\end{tcolorbox}
\end{minipage}
\caption{Topic classification prompt for the \texttt{topic\_labeling} discriminative task
         (Vietnamese original left, English translation right).}
\label{fig:prompt_topic}
\end{figure*}

\subsection{Linguistic Complexity Classification Prompt}
\label{app:disc_pattern}
The linguistic complexity classification prompt is presented in
Figure~\ref{fig:prompt_pattern}. Models are asked to identify \emph{all applicable} complexity traits for
a given idiom or proverb, returning a comma-separated list of integers
in ascending order (e.g., \texttt{1,3,5}).  An empty string is returned
when no trait applies. Exact-match scoring penalises any deviation from
the gold label, including reordering or additional characters.

\begin{figure*}[h]
\noindent
\begin{minipage}[t]{0.48\linewidth}
\begin{tcolorbox}[colback=gray!8, colframe=gray!50, title=Vietnamese]
Bạn là chuyên gia ngôn ngữ. Dưới đây là một câu thành ngữ/tục ngữ: ``\texttt{\{phrase\}}''

Nhiệm vụ của bạn là xác định câu này thuộc \textbf{những đặc điểm nào} trong danh sách dưới đây.
Kết quả trả về là danh sách số thứ tự đặc điểm, cách nhau bằng dấu phẩy (ví dụ: ``1,3,5'').
Nếu không thuộc đặc điểm nào, trả về chuỗi rỗng ``''.

\medskip
\noindent Danh sách đặc điểm:

\noindent 1. Là câu tục ngữ \textbf{không mang nghĩa bóng} (chỉ mang nghĩa đen/kỹ thuật). \\
\quad Ví dụ: \textit{Da đồng lông móc} (kinh nghiệm chọn trâu, không hàm ý sâu xa).

\noindent 2. Có \textbf{từ Hán Việt} xuất hiện. \\
\quad Ví dụ: \textit{Nhất tự vi sư, bán tự vi sư}.

\noindent 3. Có từ \textbf{ít phổ biến, ít dùng} trong ngôn ngữ hiện đại. \\
\quad Ví dụ: \textit{Am thanh cảnh vắng} (từ ``am'' nghĩa là chùa nhỏ).

\noindent 4. Thể hiện \textbf{kinh nghiệm dân gian xưa} của người Việt Nam. \\
\quad Ví dụ: \textit{Cấy thưa thừa thóc, cấy mau dốc bồ}.

\noindent 5. Mang sắc thái \textbf{mỉa mai, châm biếm hoặc tiêu cực}. \\
\quad Ví dụ: \textit{Tốt mẽ khoe màu}.

\medskip
Chỉ trả lời bằng các số đặc điểm, sắp xếp theo thứ tự tăng dần, cách nhau bằng dấu phẩy.
Không giải thích, không thêm ký tự nào khác ngoài danh sách số.
\end{tcolorbox}
\end{minipage}
\hfill
\begin{minipage}[t]{0.48\linewidth}
\begin{tcolorbox}[colback=blue!5, colframe=blue!30, title=English (translation)]
You are a language expert. Below is a Vietnamese idiom or proverb: ``\texttt{\{phrase\}}''

Your task is to identify \textbf{all applicable characteristics} from the list below.
Return a comma-separated list of characteristic numbers in ascending order (e.g., ``1,3,5'').
If none apply, return an empty string ``''.

\medskip
\noindent List of characteristics:

\noindent 1. The proverb \textbf{carries no figurative meaning} (literal/technical meaning only). \\
\quad Example: \textit{Da đồng lông móc} (buffalo selection criteria, no deeper implication).

\noindent 2. Contains \textbf{Sino-Vietnamese vocabulary}. \\
\quad Example: \textit{Nhất tự vi sư, bán tự vi sư}.

\noindent 3. Contains words \textbf{rarely used} in modern Vietnamese. \\
\quad Example: \textit{Am thanh cảnh vắng} (``am'' means a small rural temple).

\noindent 4. Reflects \textbf{traditional Vietnamese folk knowledge}. \\
\quad Example: \textit{Cấy thưa thừa thóc, cấy mau dốc bồ}.

\noindent 5. Carries a \textbf{sarcastic, ironic, or negative tone}. \\
\quad Example: \textit{Tốt mẽ khoe màu}.

\medskip
Reply only with the characteristic numbers in ascending order, comma-separated.
No explanation, no extra characters beyond the number list.
\end{tcolorbox}
\end{minipage}
\caption{Linguistic complexity classification prompt for the \texttt{pattern\_labeling} discriminative task
         (Vietnamese original left, English translation right).
         Models must predict all applicable traits; exact match is required for a correct prediction.}
\label{fig:prompt_pattern}
\end{figure*}

\section{Model and Hyperparameter Details}
\label{app:model_details}

For generation, all models use \texttt{temperature=0.7} (default) and \texttt{max\_tokens=150}.  For judging, GPT-4.1 uses \texttt{temperature=0} and \texttt{max\_tokens=10}.

\end{document}